%% file: neurips_2025.tex
\documentclass{article}

\usepackage[preprint]{neurips_2025}

\usepackage[utf8]{inputenc} 
\usepackage[T1]{fontenc}    
\usepackage{hyperref}       
\usepackage{url}            
\usepackage{booktabs}       
\usepackage{amsfonts}       
\usepackage{nicefrac}       
\usepackage{microtype}      
\usepackage{xcolor}         

\usepackage{fullpage} 
\usepackage[utf8]{inputenc} 
\usepackage[T1]{fontenc}    
\usepackage{hyperref}       
\usepackage{url}            
\usepackage{booktabs}       
\usepackage{amsfonts}       
\usepackage{nicefrac}       
\usepackage{microtype}      
\usepackage{xcolor}         
\usepackage{amsmath,amssymb}
\usepackage{algorithm}
\usepackage{algorithmic}
\usepackage{graphicx}
\usepackage{multicol}
\usepackage{tikz}
\usepackage{subcaption}
\usepackage{bbm}
\usepackage{array}
\usepackage{xspace}
\usepackage{adjustbox}

\newcommand{\benchmark}{CHARTOM\xspace}

\title{\benchmark: A Visual Theory-of-Mind Benchmark for LLMs on Misleading Charts}

\author{
Shubham Bharti$^1$
\And 
Shiyun Cheng$^1$
\And
Jihyun Rho$^1$
\And
Jianrui Zhang$^1$
\AND
Mu Cai$^{1,3}$ \thanks{Work done at UW Madison}
\And
Yong Jae Lee$^1$
\And
Martina Rau$^2$
\And Xiaojin Zhu$^1$
}

\begin{document}

\maketitle

\hspace{1cm}$^1$ University of Wisconsin-Madison \qquad $^2$ ETH Zurich \qquad $^3$ Google DeepMind 

\begin{abstract}
We introduce \benchmark, a visual theory-of-mind benchmark designed to evaluate multimodal large language models' capability to understand and reason about misleading data visualizations though charts.
\benchmark consists of carefully designed charts and associated questions that require a language model to not only correctly comprehend the factual content in the chart (the FACT question) but also judge whether the chart will be misleading to a human readers (the MIND question), a dual capability with significant societal benefits.
We detail the construction of our benchmark including its calibration on human performance and estimation of MIND ground truth called the Human Misleadingness Index.
We evaluated several leading LLMs---including GPT, Claude, Gemini, Qwen, Llama, and Llava series models---on the \benchmark dataset and found that it was challenging to all models both on FACT and MIND questions. This highlights the limitations of current LLMs and presents significant opportunity for future LLMs to improve on understanding misleading charts.
\end{abstract}

\input{sections/1_introduction}

\input{sections/1.5_related_works}
\input{sections/2_the_benchmark}

\input{sections/3_HMI}

\input{sections/4_evaluation_llms}

\input{sections/5_conclusion}

~\\
\small{
\textbf{Acknowledgments.} We thank Robert Hawkins for discussions on the theory of mind and Yea-Seul Kim and Yuhang Zhao for discussions on computer vision for chart comprehension.
This work was supported by NSF IIS 2202457.
}

\newpage

\bibliography{neurips_2025}
\bibliographystyle{plain}


\newpage
\input{sections/appendix}

\end{document}

%% file: sections/1_introduction.tex
\section{Introduction}

For AI to better assist humans, it must know not just factual truth but also how humans perceive the truth.  The two could differ for many reasons such as information asymmetry or cognitive ability.  If AI can detect the difference, it can help humans accordingly.
The ability of an individual to reason about how others think (rather than the factual truth) is known as the theory of mind~\cite{premack1978does}. 
A classic example is the Sally-Anne test~\cite{wimmer1983beliefs,baron1985does}: Sally hides a marble in a basket and then leaves the room. Anne moves the marble to a box while Sally is away.  Sally comes back.  
At this point, one can ask an observer two types of questions:
\begin{itemize}
\item \textbf{FACT} question: Where is the marble?
\item \textbf{MIND} question: Where will Sally look for the marble?
\end{itemize}
There is a recent surge of interest in AI theory of mind, where large language models take the place of the observer.
Several researchers found current AI competent at theory of mind tasks, performing near or surpassing human performance~\cite{kosinskitheory,strachan2024testing},
though some others remained skeptical~\cite{ullman2023large,sap2022neural}.
Concerned with limitations of AI theory of mind benchmarks, recent work also started to develop new benchmarks emphasizing conversations~\cite{kim2023fantom}, causality~\cite{gandhi2024understanding}, and actions~\cite{zhou2023far}.
However, these existing AI theory of mind benchmarks are all based on text comprehension.

\begin{figure}[H]
\centering

\begin{subfigure}[t]{0.30\textwidth}
  \centering
  \includegraphics[width=\linewidth]{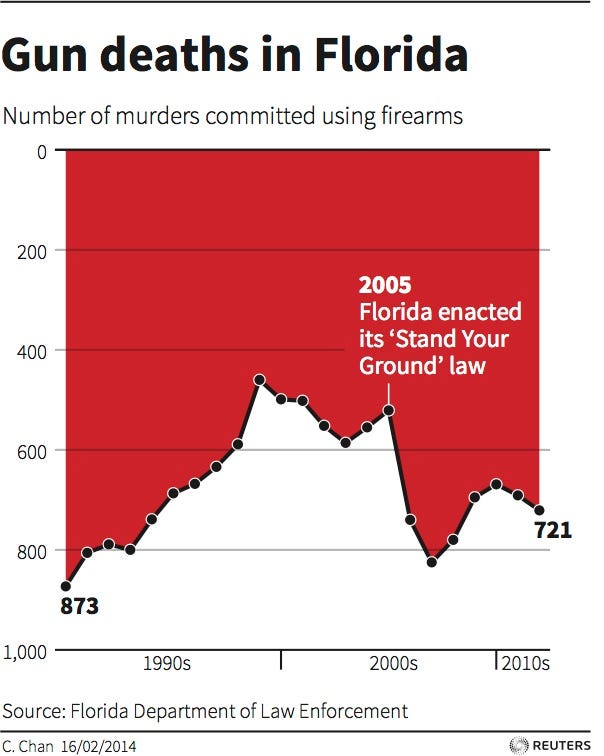}
  \caption{Reproduction of a misleading chart from~\cite{BusinessInsider}, attributed to Reuters originally}
  \label{fig:florida}
\end{subfigure}
\hfill
\begin{subfigure}[t]{0.68\textwidth}
  \centering
  \begin{subfigure}[t]{0.49\textwidth}
    \centering
    \includegraphics[width=\linewidth]{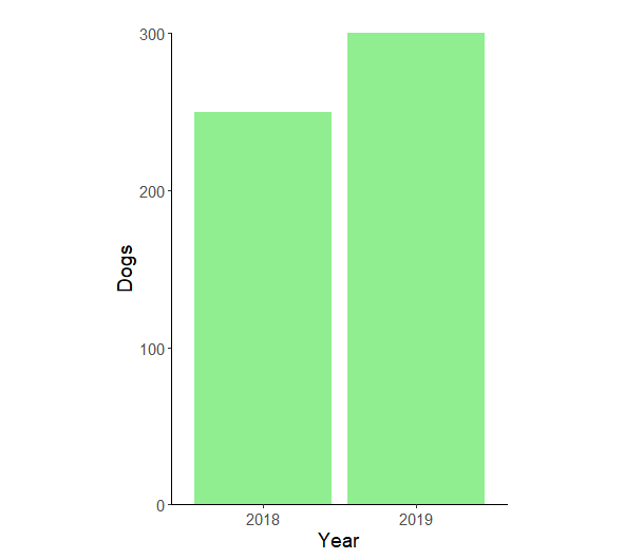}
    \caption{Non-manipulated bar chart}
    \label{fig:non_manipulated_example}
  \end{subfigure}
  \hfill
  \begin{subfigure}[t]{0.49\textwidth}
    \centering
    \includegraphics[width=\linewidth]{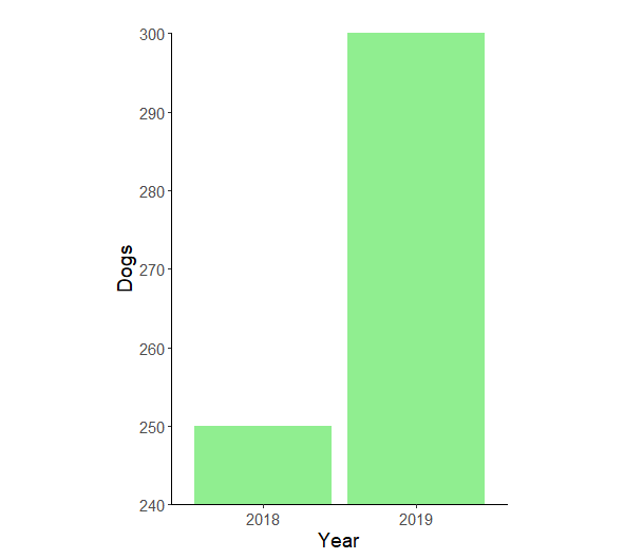}
    \caption{Manipulated bar chart}
    \label{fig:manipulated_example}
  \end{subfigure}
  \caption*{A pair of example charts in \benchmark}
  \label{fig:rightgroup}
\end{subfigure}

\caption{Example of manipulated charts in the wild and in our \benchmark dataset.}
\label{fig:example_charts}
\end{figure}

Our theory of mind benchmark tests visual perception instead. 
Specifically, our focus is on \textbf{misleading data visualizing charts}, e.g. bar charts, line graphs, pie charts, etc.
Charts are widely used for conveying quantitative information to human readers. They lend credibility to accompanying messages and hence have a strong impact on human decision-making. On the flip side, however, if charts are misleading, they can amplify the impact of misinformation~\cite{lauer2020people,pandey2015deceptive}.
Figure~\ref{fig:florida} reproduces a misleading chart whose $y$-axis is inverted, which may give the impression that enacting Florida's ``stand your ground'' law reduced gun deaths.
Misleading charts are unfortunately prevalent in many areas of our lives. For example, a review of medical advertisements found that 1/3 of charts were misleading~\cite{cooper2003quantity}. A review of charts in Korean news about the COVID-19 pandemic showed that about 30\% of bar charts and 44\% of pictorial charts were misleading~\cite{kwon2021graphs}. A U.S.-based analysis of visuals that had been identified as misleading by human fact-checkers showed that more than half of these visuals were used as evidence for fake news about COVID-19~\cite{brennen2021beyond}. 
Because journalists often use charts to make narratives appear objective and to engage viewers emotionally~\cite{ryan2016visual}, misleading charts can make an inaccurate story more deceptive~\cite{diakopoulos2018ethics,west2021misinformation}.

As individuals increasingly rely on digital media to inform their opinions and decision-making processes, it is essential to ensure that the visual representations of data they encounter are accurate and trustworthy.  Studying misleading charts can lead to AI with the ability to automatically detect them and alert readers, thus empowering people to make well-informed decisions based on reliable information, consequently promoting a more informed and discerning society.

However, we point out a significant shortcoming in chart comprehension: many misleading charts in fact contain the correct data.  Take Figure~\ref{fig:florida} for example.  Despite the unusual flipped $y$-axis, a meticulous reader (or AI) can indeed recover all the correct (year, deaths) data points from the chart.  Therefore, ironically, a perfect chart comprehension system is not directly suited for judging how misleading a chart may be to humans.
This is the same distinction between factual and theory of mind questions. We can ask:
\begin{itemize}
    \item A \textbf{FACT} question: \emph{What data is objectively presented in the chart?}
    \item A \textbf{MIND} question: \emph{What will a typical human reader perceive from the chart?}
\end{itemize}
Current chart comprehension systems aim to solve FACT questions. To the best of our knowledge, the present paper is the first to address the MIND question in charts instead.

\textbf{Our Contributions}: Towards this end, we make the following contributions:
\begin{enumerate}
    \item We create a visual theory-of-mind benchmark \benchmark (for CHARt Theory Of Mind) consisting of specially designed charts.  The benchmark covers common variety of chart types including line, bar, scatter, map, and pie charts as shown in Table~\ref{tab:graph_types_scores}
    \item We designed two types of questions for each chart: 1.) a FACT question that tests the understanding of factual information present in charts, and, 2.) a MIND question that tests the understanding of ``how much'' and ``why'' a chart is misleading to humans.
    \item The key to the benchmark is MIND ground truth called HMI (Section~\ref{sec:MIND_groundtruth}) that quantifies the ``degree of misleadingness to humans'', of each chart that has been obtained via human experiments. We also provide ground truth reasoning that captures why humans are misled.
    \item We test the performance of some of the leading LLMs in Claude, Gemini, GPT, Llama, Llava, and Qwen series and report our findings in Section~\ref{sec:evaluation}. We observe that most LLMs perform poorly not only on MIND but also on FACT questions indicating their failure to understand manipulated charts properly. This also suggests a big scope for improvement in future LLMs to understand misleading charts. 
\end{enumerate}

%% file: sections/1.5_related_works.tex
\section{Related Works}

Recent works have extensively explored the capability of modern large language models (LLMs) to understand and reason about theory of mind(TOM) problems, focusing on narrative or conversational tasks involving belief attribution and social cognition \cite{kosinskitheory, street2024motom, chen2024tombench}. Prior standard TOM benchmarks typically consist of tasks such as false-belief reasoning \cite{kosinskitheory}, second-order inference \cite{street2024motom}, and social reasoning tasks like faux pas detection, hint interpretation, irony, and sarcasm understanding \cite{strachan2024human} all based on textual data. These studies have demonstrated that advanced LLMs, particularly GPT-4, approach or even surpass human-level performance on various TOM tasks \cite{chen2024tombench, strachan2024human}. However, despite the promising results, LLMs often exhibit brittleness, and struggle with implicitly conveyed mental states \cite{shapira2023faux, chen2024tombench}. Unlike these works, we focus on a very different visual theory of mind problem for understanding charts that mislead humans into drawing the wrong conclusion about presented data.

Existing computer vision research in charts has largely focused on chart comprehension.
Classic approaches often employ an intermediate step of chart-to-data parsing, then use the data for downstream tasks such as visual question-answering, reasoning, and summarization.
To this end,~\cite{visual2019,chartvi2022} applied deep learning-based object detection methods to extract data, while others~\cite{bar2015,scatteract2017,ivolver2016,chartsense2017,revision2011} applied semi-automated off-the-self OCR text recognition systems~\cite{datathief,tesseract2007}.
Newer LLM-based approaches tend to be end-to-end, such as training the model on extensive pre-training datasets~\cite{unichart2023,chartllama2023,chartassisstant2024,zhang2024tinychart}, few-shot tuning methods which fine-tune pre-trained models with a small number of task-specific examples~\cite{do2023llms,liu2023deplot,kim2024simplot}, and even zero-shot learning where the model makes predictions on tasks it has not been explicitly trained on~\cite{wu2024evaluating,islam2024large}.  The newer approaches are performant, though occasionally still suffer from hallucination in data extraction and reasoning~\cite{huang2024pixels,advances2024}. Nevertheless, all the prior works have mainly focused on standard chart comprehension which is very different from our theory of mind understanding of misleading charts.

%% file: sections/2_the_benchmark.tex
\section{The \benchmark Benchmark}
In this section, we present our visual theory of mind benchmark \benchmark.
We first detail our design of charts to accommodate both FACT and MIND type of questions in the benchmark, then explain how we obtain the MIND ground truth with human experiments.

\subsection{Design of the \benchmark Benchmark}

To explain our \benchmark benchmark, let us start with an example.

\textbf{An example.} 
The pair of charts in Figure~\ref{fig:example_charts} share the same underlying data: (year=2018, dogs=250) and (year=2019, dogs=300).
The left chart has a standard design which we call ``non-manipulated'', while in the right chart, we intentionally planted a potentially misleading manipulation: we truncated the $y$-axis range so it starts at 240 instead of 0.
This is to create a perceptual effect: the 2019 bar now appears much taller than the 2018 bar.
For each of the two charts, we ask the following FACT and MIND questions.
\begin{itemize}
\item The \textbf{FACT} question is a chart comprehension question: 
\begin{quote}
The following graph shows the number of dogs adopted. The dogs adopted in 2018 eat 1 million bags of dog food in their lifetimes. How much do the dogs adopted in 2019 eat in their lifetimes?
\end{quote}

\item The \textbf{MIND} question is a human performance question:
\begin{quote}
Here is a chart we will present to typical university students and ask them the following question:

[THE FACT QUESTION ABOVE]

What fraction of typical university students do you predict will be misled by the chart when answering the question? First, give your prediction as a decimal number between 0 and 1, then justify your prediction in words.
\end{quote}
\end{itemize}

\textbf{The \benchmark  benchmark.}
Our entire \benchmark benchmark follows the same design principle.  
The benchmark consists of 112 charts, covering 
five prevalent chart types: line, bar, pie, scatter plot, and map. 
The charts are manually designed to be visually tidy.
We present selected charts from the benchmark in Table~\ref{tab:graph_types_scores} and Appendix A.

The 112 charts come in 56 pairs (one pair is already shown in Figure~\ref{fig:example_charts}). 
Each pair has a non-manipulated (suffix \_1) and a manipulated (suffix \_2) version, the latter incorporates one of many visualization fallacies suggested by the psychology literature, see the `manipulation' column in Table~\ref{tab:graph_types_scores}.
The manipulated version is intended to mislead casual human readers.
Each pair of charts also comes with a FACT question and a MIND question.
The FACT question is typically a simple word problem based on chart comprehension, see examples in Appendix A.
There are three types of FACT questions: multiple-choice, free text entry, and ranking. 
Importantly, the true underlying data can still be recovered from either version if one is careful.
This means that the FACT question has the same answer on both versions.
The MIND question has the same format as shown earlier for all charts.

To summarize, the \benchmark benchmark consists of:
\begin{itemize}
\item 112 charts: 56 pairs of non-manipulated and manipulated charts.
\item Each chart comes with a FACT question and an answer key.
\item Each chart comes with a MIND question, and the answer key is the Human Misleadingness Index (HMI, see Section~\ref{sec:MIND_groundtruth}) obtained from human experiments. 
\end{itemize}

\begin{table}[ht]
\centering
\begin{tabular}{c|l|c|c|l|c}
\toprule
chart type & answer type & group ID &  HMI(non-manip) & manipulation type & HMI(manip) \\
\midrule
line & free text & G1 & 0.51 & truncated y-axis & 0.68 \\
line & free text & G2 & 0.76 & compressed y-axis & 0.88 \\
line & multiple choice & G3 & 0.03 & inverted y-axis & 0.83 \\
line & multiple choice & G4 & 0.05 & inconsistent x-axis & 0.92 \\
line & multiple choice & G9 & 0.19 & dual axes & 0.84 \\
\midrule
bar & free text & G7 & 0.31 & truncated y-axis & 0.53 \\
bar & free text & G8 & 0.76 & compressed y-axis & 0.69 \\
bar & free text & G10 & 0.19 & pictorial bars & 0.27 \\
bar & multiple choice & G12 & 0.11 & 3D effect & 0.48 \\
\midrule
scatter & multiple choice & G5 & 0.02 & logarithmic y-axis & 0.63 \\
scatter & multiple choice & G6 & 0.02 & inverted x-axis & 0.73 \\
\midrule
map & multiple choice & G15 & 0.02 & inverted color scale & 0.15 \\
\midrule
pie & rank & G13 & 0.25 & 3D effect & 0.79 \\
pie & rank & G14 & 0.30 & 3D effect + pop out & 0.79 \\
\bottomrule
\end{tabular}
\vspace{0.1cm}
\caption{MIND ground truth answers (HMI) score averaged across various manipulation groups in the \benchmark benchmark.  Each row is for non-manipulated and manipulated version of charts within a manipulation group. HMI stands for Human Misleadingness Index, a number between 0 and 1 where a larger value suggests the chart is more misleading to humans.}
\label{tab:graph_types_scores}
\end{table}

%% file: sections/3_HMI.tex
\section{Obtaining Ground Truth Answers}
\subsection{Ground Truth for FACT Questions}
\label{sec:FACT_evaluation}
The FACT questions are based on ground truth data. Their ground truth answers are easily computable.
Recall both non-manipulated and manipulated versions by design have the same underlying data, and that data can be exactly recovered regardless of visual manipulation.
Therefore, a perfect human or AI with proper reasoning can always obtain the same correct FACT answer on both chart versions.
For example, in Figure~\ref{fig:example_charts} the correct FACT answer is $1 \times \frac{300}{250}=1.2$ million bags.

Specifically, on FACT questions we suggest using accuracy, defined as the number of correct answers over the number of FACT questions.
To define correctness, we consider the three types of FACT questions separately:
\begin{enumerate}
\item Multiple choice questions.
An answer is correct if it precisely matches the answer key.

\item Ranking questions.
A ranking answer is correct only if all items are ranked correctly. 

\item 
Numerical (free text) questions. 
We provide a correct answer key assuming perfect perception.
But an exact match can be too strong an assumption for either humans or LLMs when they can enter any free-text number.
So, we suggest using a $\pm 10\%$ tolerance interval: if the answer key is $a$, then an answer in the interval $[0.9a, 1.1a]$ is deemed correct.

\end{enumerate}

\subsection{Ground Truth for MIND Questions}
\label{sec:MIND_groundtruth}
In contrast, the MIND question is trickier to judge.  On the non-misleading chart in Figure~\ref{fig:example_charts}(left) where the design is standard, one might still argue that the chart is not perfect and could be a little bit misleading to humans: the numbers are too small, there is no grid guidelines to help visual alignment, etc.
Conversely, on the misleading chart (right) even if we intentionally planted the truncated $y$-axis because we \emph{presumed} it to be misleading based on existing psychology literature (e.g.~\cite{lo2022misinformed}), we do not know \emph{a priori} the actual effect on human readers.
To avoid these difficulties we propose a metric called Human Misleadingness Index as defined next.

\textbf{Human Misleadingness Index (HMI)} of a chart is a number between 0 and 1 that quantifies the fraction of the human population that is misled by the chart.

The correct answer to a MIND question is the HMI value. To obtain HMI we conducted human experiments, which is expensive and time-consuming but is the only proper method.
Our human experiment involves 68 university students with IRB approval. 
We did not directly ask human participants the MIND question because we do not anticipate people to be able to give reliable answers---after all, many were themselves misled by some of our charts.
Instead, we asked them the FACT question, which is designed so that users who are misled by the chart are likely to give wrong answers. 
We randomized the order of non-manipulated and manipulated charts in each pair when presenting them to human participants to minimize the order effect.
We estimate \emph{the HMI of a chart by the percent of our study population who did not give a correct answer on the FACT question}.

We argue that HMI is a surrogate measure for how misleading the chart is.
This definition is motivated by two factors:
\begin{enumerate}
\item
The arithmetic involved in solving the FACT questions is straightforward for the education level of our participants.  Indeed on most non-manipulated charts, the HMI is small.
\item
On many manipulated charts the HMI is higher compared to their non-manipulated chart counterpart.
Since the only difference between the pair is the planted manipulation in data visualization, we can attribute the increase in HMI to the manipulation being more misleading to humans.
\end{enumerate}

\begin{figure}
\begin{center}
\includegraphics[width=0.5\textwidth]{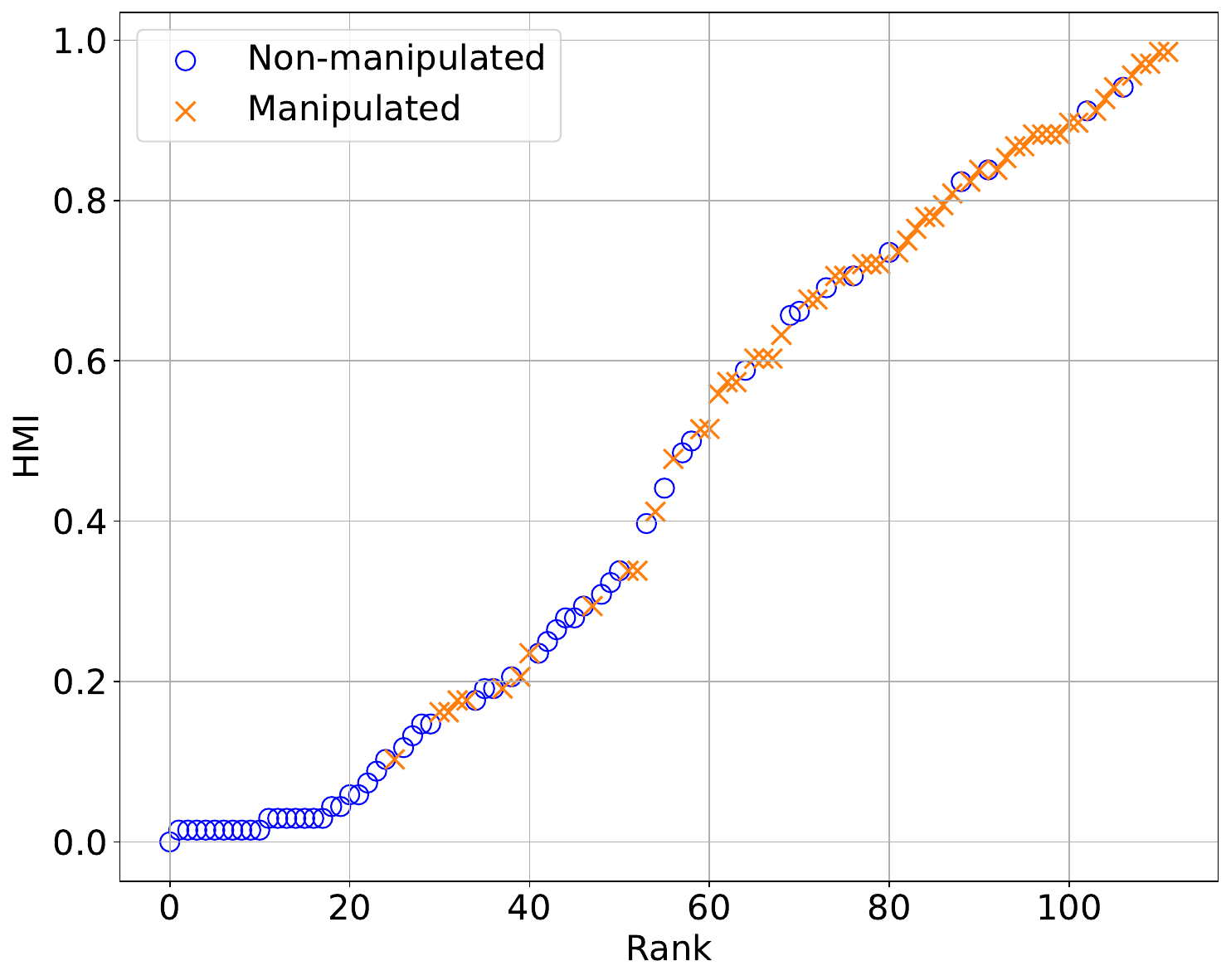}
\caption{The HMI values of the 112 charts in the \benchmark benchmark, sorted.}
\label{fig:HMI}
\end{center}
\end{figure}

Figure~\ref{fig:HMI} shows the HMI values in the \benchmark benchmark.
These values are sorted from small (not misleading to humans) to large (very misleading).
In addition, each HMI value is color-coded to indicate whether the chart is the non-manipulated version (blue) or the manipulated version (orange).
Generally speaking, the non-manipulated charts have small HMI, and the manipulated ones have large HMI.
But this depends on the type of manipulation.
As the HMI values in Table~\ref{tab:graph_types_scores} indicate, some manipulations greatly misled people: inconsistent x-axis, inverted y-axis, dual axes plots, 3D effect, etc.
While interestingly, some other manipulations did not confuse people much: compressed y-axis and pictorial bars are benign. 
Such differences are of interest to cognitive science.
Important details of our human experiments are presented in the appendix.

%% file: sections/4_evaluation_llms.tex
\section{Benchmarking Leading LLMs}\label{sec:evaluation}
We evaluated leading LLMs including Claude-3.7-Sonnet (claude-3.7)~\cite{anthropic2025sonnet3.7}, Claude-3.5-Sonnet (claude-3.5)~\cite{anthropic2024Claude3family}, Gemini-1.5-Pro (gemini-1.5)~\cite{Geminiteam2024Gemini15unlockingmultimodal}, Gemini-2.5-Flash (gemini-2.5)~\cite{google2025gemini2.5}, GPT-4o (gpt-4o)~\cite{openai2024gpt4ocard}, OpenAI o1 (gpt-o1)~\cite{openai2024openaio1card}, Llama-3.2-90B (llama-3)~\cite{grattafiori2024Llama3herdmodels}, LLaVA-OneVision-72B (llava)~\cite{li2024LLaVAonevisioneasyvisualtask}, Qwen2-VL-72B (qwen-2)~\cite{yang2024Qwen2technicalreport}, Qwen-2.5-VL-72B (qwen-2.5)~\cite{bai2025qwen25vltechnicalreport} on both FACT and MIND questions and present our findings below.
When evaluating LLMs, we suggest the following criteria:
\begin{itemize}
    \item On FACT questions, we suggest reporting accuracy based on the evaluation in section \ref{sec:FACT_evaluation}.
    \item On the numerical part of the MIND questions, we suggest reporting mean absolute error: $\frac{1}{ n}\sum_{i=1}^n |m_i - a_i|$ where $n$ is the number of MIND questions, $m_i$ is an LLM's answer on the $i$th MIND question, and $a_i$ is the HMI answer key to that MIND question.
    \item On the justification part of the MIND questions, we suggest qualitative comparisons to the ``chart manipulations'' file provided in the benchmark.  That file records the single manipulation we planted to the manipulated (suffix \_2) version of each chart pair.  If a manipulated chart has a high HMI value, a capable LLM should identify the corresponding manipulation as an important cause of misleadingness.
\end{itemize}

\begin{figure}
\begin{center}
\includegraphics[width=0.8\textwidth]{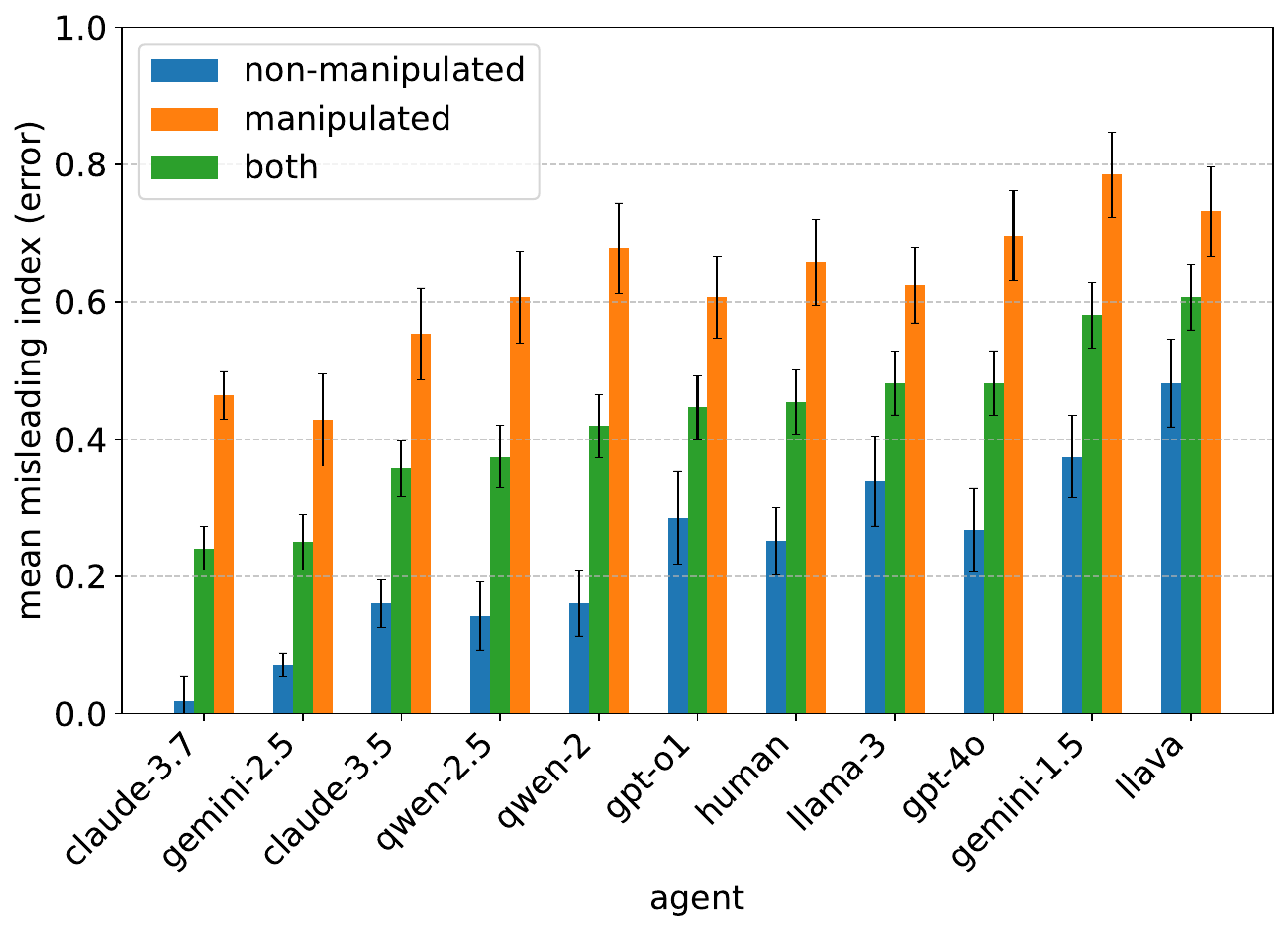}
\caption{The error rate of human participants and LLMs on FACT questions}
\label{fig:original_vs_manipulated_per_llm}
\end{center}
\end{figure}
The following are our main findings.

\subsection{Like Humans, all LLMs are Fooled by Manipulations}
We asked LLMs to answer FACT questions on both non-manipulated and manipulated charts and record their response. We then evaluate their performance based on the metric suggested above. Figure~\ref{fig:original_vs_manipulated_per_llm} shows the error rate of each LLM separately on non-manipulated charts (blue), manipulated charts (orange), and all charts (green). We make the following observations:

\begin{enumerate}
    
    \item \textbf{Top LLMs show near or superhuman performance on FACT questions.} We observe that top LLMs like claude-3.7, gemini-2.5, qwen-2.5, and gpt-4o performed better or very close to humans (on average) while others like llama-3, gemini-1.5, and llava have shown relatively poorer performance.

    \item \textbf{LLMs perform significantly worse on manipulated charts as compared to non-manipulated ones.} This is evident from the difference in height of blue and orange bars for each LLM in Figure~\ref{fig:original_vs_manipulated_per_llm}.
    We performed a $\mathcal X^2$ test that achieved a p-value $\leq0.05$ for all LLMs.
    This suggests that, like humans, LLMs are fooled by manipulations and there is big scope for improvements, particularly on manipulated charts.

    Furthermore, Some LLMs like qwen-2.5, gpt-o1, llama-3, gpt-4o fail to even understand non-manipulated charts. This suggests a good scope for future improvements.
\end{enumerate}

\subsection{Certain Manipulations are More Misleading Than Others}

\begin{figure}[h]
    \centering
    \includegraphics[width=\textwidth]{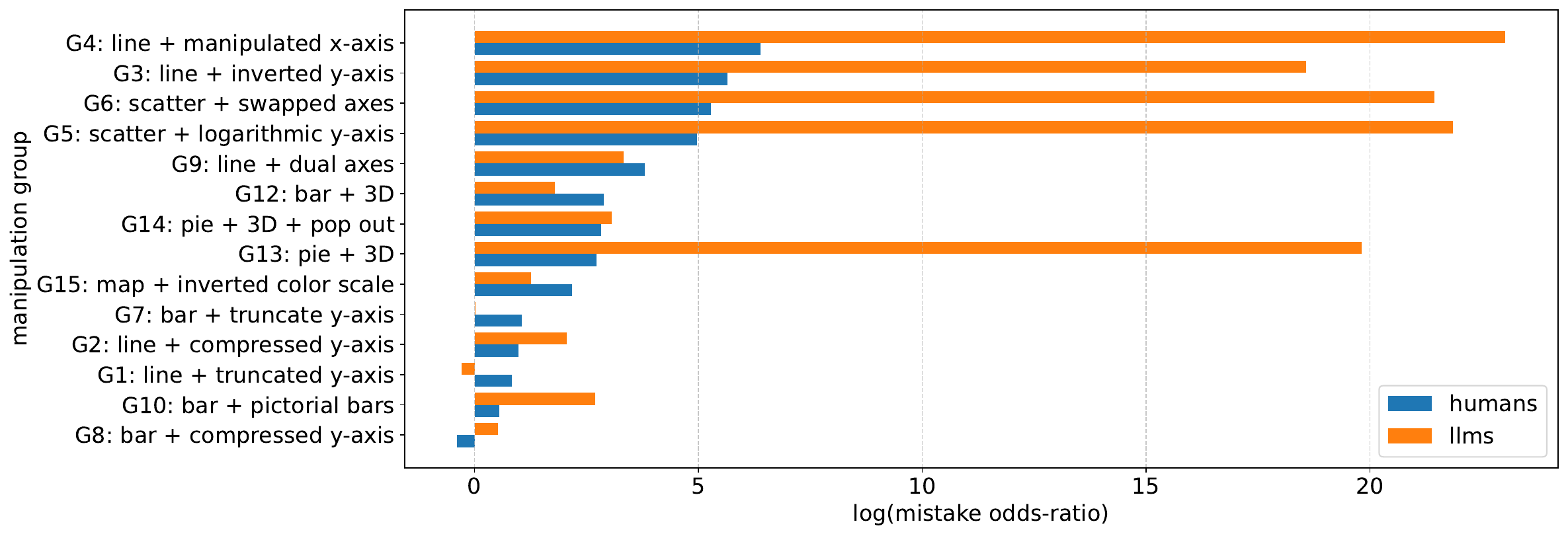}
    \caption{Log odds-ratio of making mistakes by humans and LLMs.}
    \label{fig:odds_ratio}
\end{figure}

Different manipulations have different ``degree'' by which they mislead a population of humans or LLMs. We characterize this ``degree'' by log odds ratio of making mistakes by a population on manipulated charts when compared to non-manipulated ones. 

\paragraph{Log odds ratio}: The log odds ratio characterizes the degree of misleadingness of a manipulation is computed as follows: 
\begin{equation}\label{eqn:log_odds_ratio}
\log(OR) = \log\left(\frac{\hat p_m/(1-\hat p_m)}{\hat p_n/(1-\hat p_n)}\right)
\end{equation}
where, $\hat p_m, \hat p_n$ are the maximum likelihood estimates of making mistakes on manipulated and non-manipulated charts, respectively, by a population of agents. A higher odds ratio suggests that manipulation is more misleading.

We report the log odds ratio for each manipulation group by both humans and LLMs in Figure~\ref{fig:odds_ratio}. To estimate $\hat{p}_m$ and $\hat{p}_n$, we fit separate mixed-effects logistic regression models on human and LLM responses. For both human and LLM populations, the model considers group id and manipulation type as fixed effects, and agent id and question id within each group id as random effects. The estimated marginal means from these models were used to obtain $\hat{p}_m$ and $\hat{p}_n$ values which were then used for computing the log-odds ratio in equation~\ref{eqn:log_odds_ratio}.

We make the following observations from Figure~\ref{fig:odds_ratio}:
\begin{enumerate}
    \item \textbf{Manipulations that effectively mislead humans tend to also deceive LLMs}: In Figure~\ref{fig:odds_ratio}, we show different manipulations in decreasing order of their ``degree" of misleadingness to humans along with corresponding log odds ratio for both humans and LLMs. We observe a moderate to strong correlation (Kendall's Tau score of 0.60) between the rankings of 14 manipulations between humans and LLMs. Furthermore, we note that, 
    \begin{itemize}
        \item Top manipulation like line+manipulated x-axis, line+inverted y-axis, scatter + swapped axes,  scatter+logarithmic y-axis are significantly misleading to both humans and LLMs.
        \item Bottom manipulations like bar+truncated y-axis, bar+compressed y-axis do not mislead humans. However, some of them like bar+pictorial bar and line+compressed y-axis, pie+3D still mislead LLMs though less strongly.
    \end{itemize}
    \item \textbf{The two groups have significantly different scales of log odds ratio}: The log odds ratio of LLMs on top manipulations in Figure~\ref{fig:odds_ratio} is on the order of $10^{20}$, compared to that of humans which is roughly on the order of $10^5$. This shows that when LLMs are misled, they are misled much more significantly. This suggests different modes of failure.
\end{enumerate}

\subsection{LLMs Failed to Estimate Chart Misleadingness to Humans (HMI)}

We asked each LLM the MIND question (what fraction of humans will be misled) on each chart.
Table~\ref{tab:MAE} reports the mean and standard deviation of the absolute error between the LLM-predicted and ground truth HMI scores.

\begin{table}[h]
\centering
\begin{adjustbox}{max width=\textwidth}
\begin{tabular}{lcccccccccc}
\toprule
 & gemini-2.5 & qwen-2 & qwen-2.5 & llava & gpt-o1 & gpt-4o & gemini-1.5 & claude-3.7 & claude-3.5 & llama-3 \\
\midrule
MAE & 0.24 & 0.29 & 0.30 & 0.32 & 0.32 & 0.35 & 0.37 & 0.37 & 0.40 & 0.46 \\
$\hspace*{0.25cm} \sigma$   & 0.16 & 0.18 & 0.22 & 0.18 & 0.20 & 0.22 & 0.24 & 0.23 & 0.22 & 0.27 \\
\bottomrule
\end{tabular}
\end{adjustbox}
\caption{MAE of predicting HMI(MIND answer) by LLMs.}
\label{tab:MAE}
\end{table}

\begin{figure}[htb]
\begin{center}
\includegraphics[width=\textwidth]{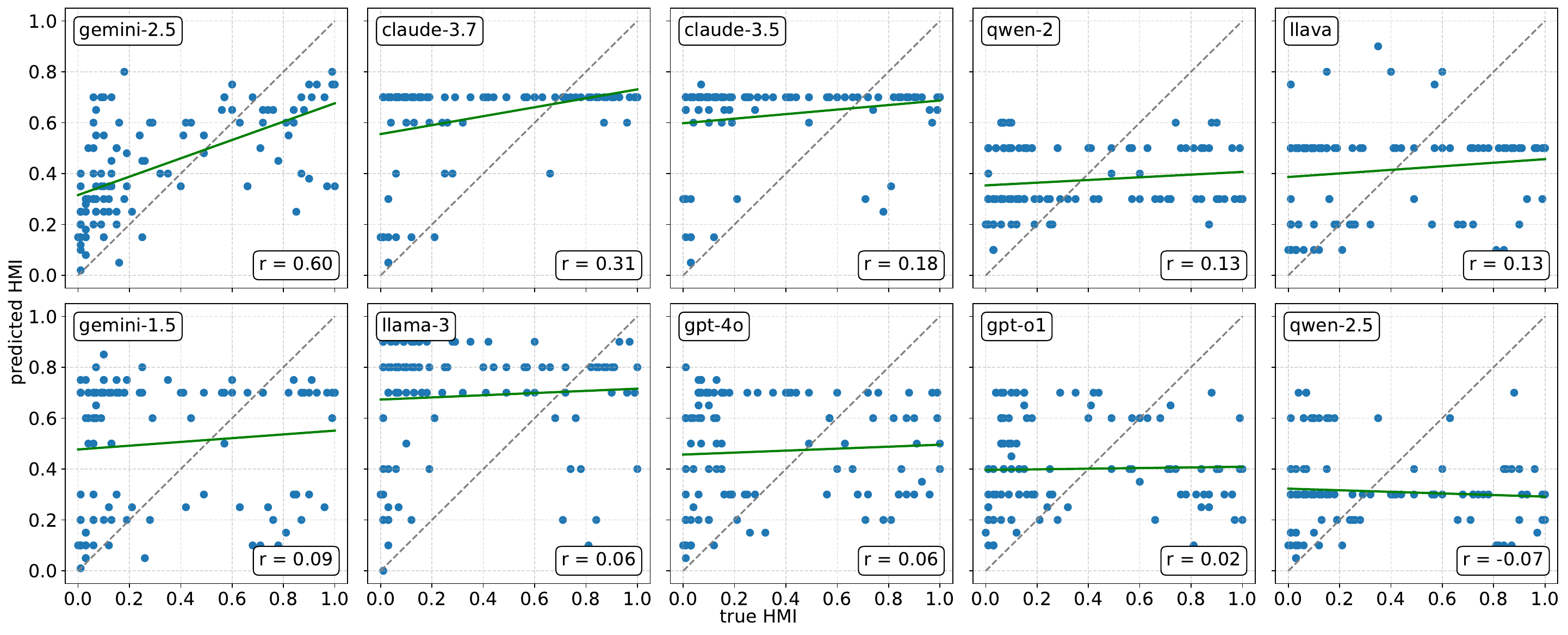}
\caption{LLMs prediction vs ground truth on MIND questions.}
\label{fig:tom_per_llm_scatter_plot}
\end{center}
\end{figure}

We also show the descriptive performance of each LLM on the MIND question through scatter plots in Figure~\ref{fig:tom_per_llm_scatter_plot}. The x-axis represents the ground truth HMI of all charts sorted in increasing order. A perfect prediction should align the dots along $y=x$ lines. We fit a linear curve on the prediction made by LLMs and show it by green line. We also show Pearson's degree of correlation between actual HMI and predicted HMI on lower end corner for each LLM.

We make the following observations based on these results:
\begin{enumerate}
    \item \textbf{LLMs lack visual theory of mind on charts}: This is evident from the large MAE values for each LLM which suggests that they cannot reliably predict the extent to which humans are misled by various chart manipulations.
    
    \item \textbf{LLMs show limited discriminative power between high and how HMI charts}: Most LLMs except gemini-2.5 show almost a flat trend(as shown by the green line in Figure~\ref{fig:tom_per_llm_scatter_plot}) in predicting HMI. This indicates that they can't effectively differentiate between low and high HMI charts. However, gemini-2.5 shows some linear trend with a Person's correlation of 0.6 suggesting that it can predict misleading aspects in charts much better. 
    
    As a corollary, we remark that LLMs over or under-estimate HMI of ground truth low or high HMI charts respectively. This trend is also evident from the per manipulation arrow plot presented in Figure~\ref{fig:tom_per_llm_by_manipulation_type} in the appendix.
\end{enumerate}

%% file: sections/5_conclusion.tex
\section{Impacts \& Limitations}

Our work takes an important step towards understanding how LLMs understand and interpret misleading charts which is an area of great societal value. By releasing this novel benchmark, we lay the foundation for developing future LLMs that can not only recognize misleading charts but can also serve as useful assistants to warn and educate human users against misinformation to curb them.

A primary limitation of our benchmark is the relatively small size of our data, which is due in part to the high cost of conducting human experiments in a controlled setting. Additionally, our dataset mainly focuses on real-world charts that can be programmatically generated. As a result, it does not capture the full diversity of chart formats that may appear in the wild, including those created by non-standard means. 

\section{Conclusion}
We presented a new theory-of-mind benchmark called \benchmark for LLMs based on comprehension of misleading charts. Our benchmark contained two types of questions, one that tested understanding of LLMs on factual content present in charts, and, the other tested understanding of LLM on how much and why a chart is misleading to the human mind. 

We evaluated the performance of leading LLMs on both FACT and MIND questions and observed that current LLMs not only possess a poor understanding of the degree of misleadingness of charts to humans but also fail to answer factual questions on manipulated charts thereby suggesting a big scope of improvement in future LLMs to understand and help humans with misleading charts.

%% file: sections/appendix.tex
\section*{Appendix A: Additional Results}
\renewcommand{\thesubsection}{\Alph{subsection}}

\subsection{FACT Results}
\begin{figure}
    \begin{center}
    \includegraphics[width=\textwidth]{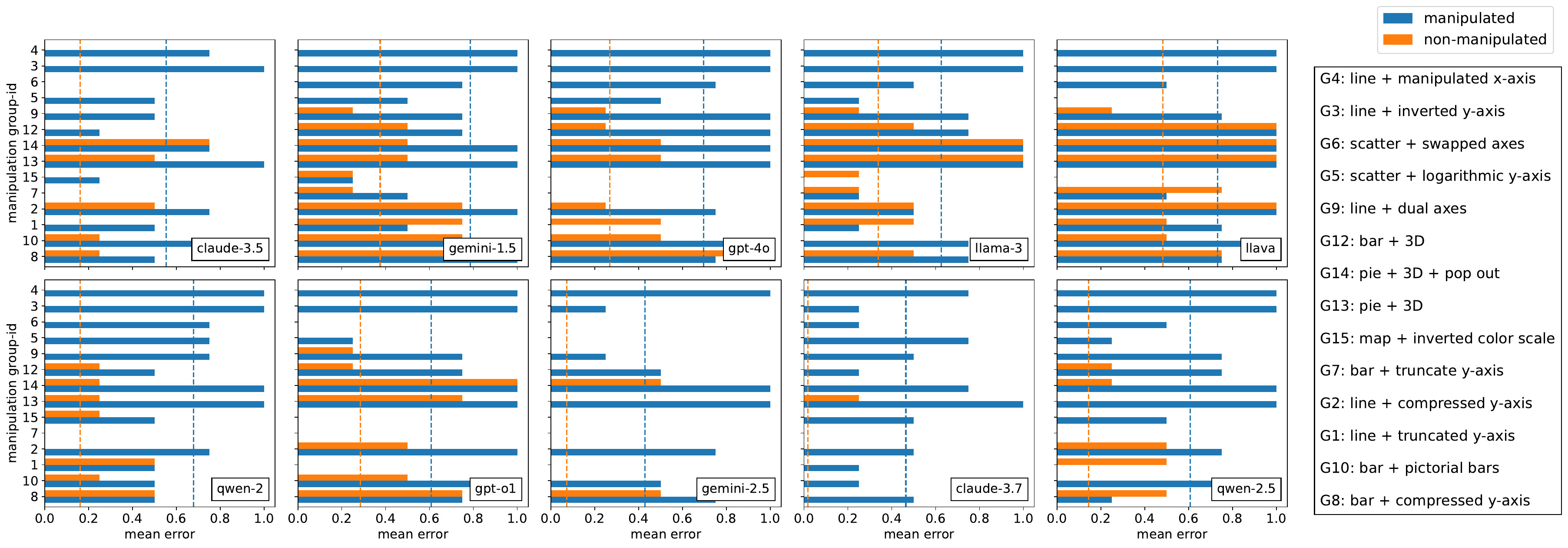}
    \caption{Performance of LLMs on FACT questions for each manipulation group.}
    \label{fig:original_vs_manipulated_per_manipulation_group_per_llm}
    \end{center}
\end{figure}

We also analyze the performance of each LLM on different manipulation groups, report their performance in Figure~\ref{fig:original_vs_manipulated_per_manipulation_group_per_llm} and make the following observations:
\begin{enumerate}
    \item For manipulations like line + inverted y-axis and manipulated x-axis, all LLMs perform perfect on non-manipulated versions but are totally fooled on manipulated versions.
    \item Some models like Llama, and Llava exhibit poor performance on even non-manipulated versions containing pie, line and bar charts highlighting that they lack the understanding of even standard charts and questions of these groups.
    \item Surprisingly, most models perform poorly on both non-manipulated and non-manipulated versions of bar + compressed y-axis and pie chart with Claude performing the best among on prior and Qwen performing best on the later one. 
\end{enumerate}

\subsection{MIND Results}

\begin{figure}
\begin{center}
\includegraphics[width=\textwidth]{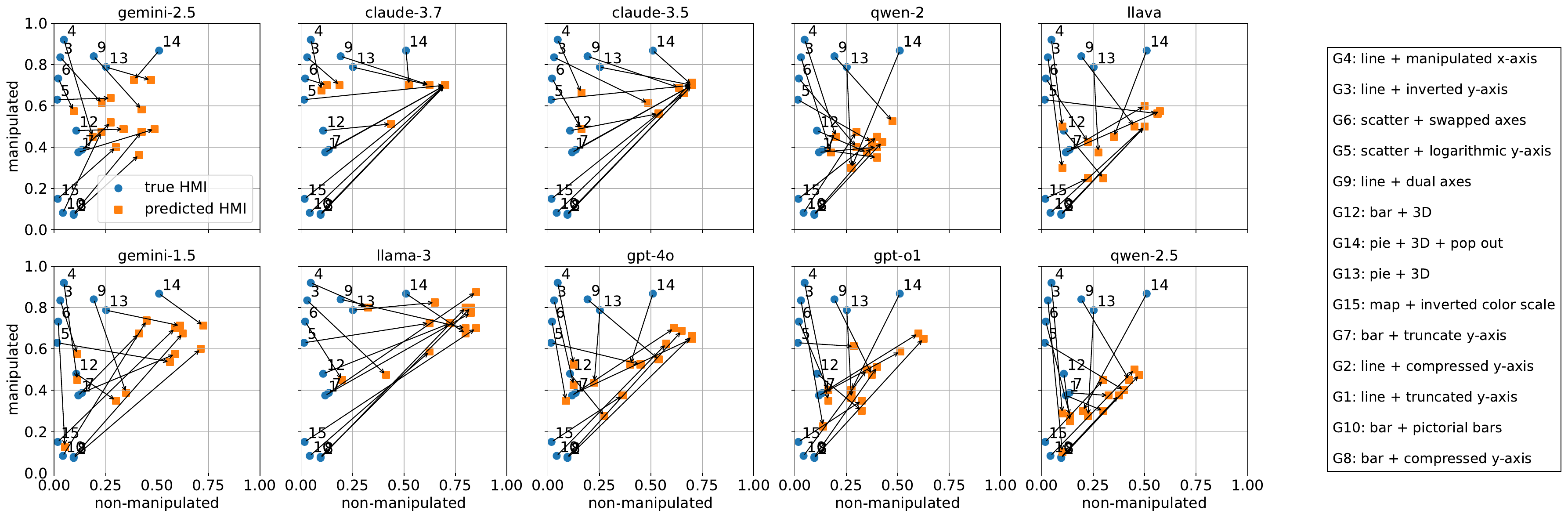}
\caption{Performance of different LLMs on MIND task.}
\label{fig:tom_per_llm_by_manipulation_type}
\end{center}
\end{figure}

We also show finer predictions made by LLMs on each manipulation group in Figure~\ref{fig:tom_per_llm_by_manipulation_type}. Each black dot represents the ground truth HMI for the manipulated version(on the y-axis) and the non-manipulated version(on the x-axis). The corresponding HMI prediction made by LLM is shown with a colored dot joined by an arrow. We make the following observations:

\begin{enumerate}
    \item For manipulation groups that are low misleading(black dots 2,8,10,15 in the bottom left of Figure~\ref{fig:tom_per_llm_by_manipulation_type}), all LLMs consistently overestimate the misleadingness in both non-manipulated and manipulated version of the group. This is evident from the corresponding arrows getting stretched towards the top right.
    \item For highly misleading manipulation groups (black dots 3,4,5,6 in the top left of Figure~\ref{fig:tom_per_llm_by_manipulation_type}), all LLMs consistently underestimate the misleadingness in manipulated version and overestimate misleadingness in non-manipulated version. This is evident from the corresponding arrows getting stretched towards the bottom right. 
\end{enumerate}

\subsection{Logistic Regression Models for Estimating $\hat{p}$}

To estimate the probabilities of making mistakes ($\hat{p}_m$, $\hat{p}_n$) for both LLMs and humans, we fit separate mixed-effects logistic regression models using the \texttt{lme4} package in R.

\paragraph{LLM Responses.}  
The following model was used to estimate the probability of making error for large language models:
\begin{verbatim}
glmer(error ~ group_id:manipulation_type + 
      (1 | llm) + (1 | group_id:question_id), 
      data = data, family = "binomial")
\end{verbatim}

Note that we have used group id and manipulation type as fixed effects and llm identity and question id of different examples within a particular group as mixed effect.

\paragraph{Human Responses.}  
A structurally similar model was used for human responses, replacing \texttt{llm} with student id:
\begin{verbatim}
glmer(error ~ group_id:manipulation_type + 
      (1 | student_id) + (1 | group_id:question_id), 
      data = human_data, family = "binomial")
\end{verbatim}

\newpage
\section{Details on Human Experiment}

To compute the Human Misleadingness Index (HMI), we collected data from 78 undergraduate participants at a Midwestern U.S. university. Participants completed a within-subject experiment in which they viewed 112 visualizations—56 with manipulated visualizations and 56 corresponding non-manipulated counterparts—representing 14 types of manipulations (e.g., truncated y-axis, inverted y-axis, dual axes). For each visualization, participants answered a question requiring data interpretation, such as comparing values or identifying trends. The study began with informed consent and an instructional video simulating real-world scenarios (e.g., scrolling through graphs on social media), after which participants completed the visualization interpretation tasks at their own pace. Sanity-check questions were included to filter inattentive responses, and 10 participants were excluded based on this criterion. The final dataset from 68 participants was used to compute accuracy-based scores, quantifying the deceptive impact of each manipulation type as perceived by human viewers.

A screenshot of chart with questions presented to human subjects are given below(chart id is provided here just for reference):

\begin{figure}[htbp]
    \centering
    \includegraphics[width=0.48\textwidth]{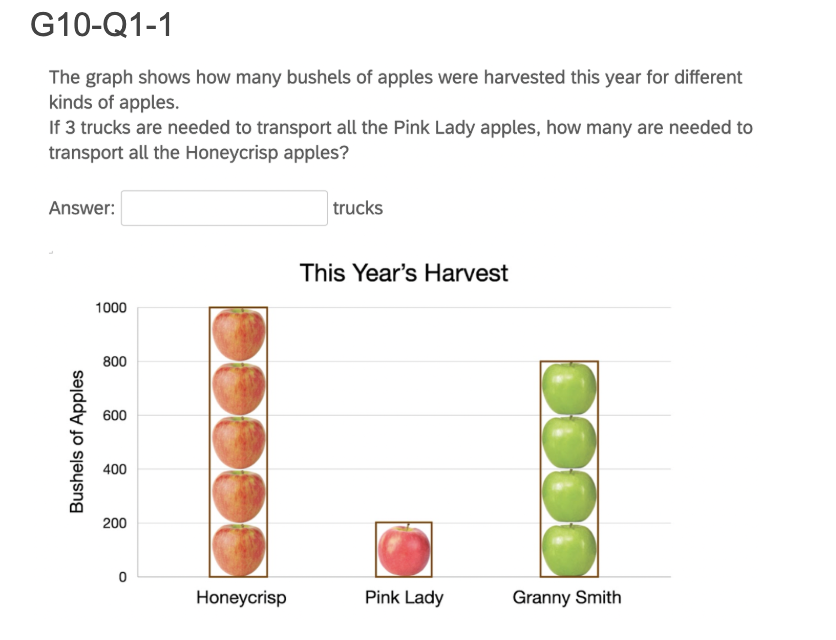} 
    \includegraphics[width=0.48\textwidth]{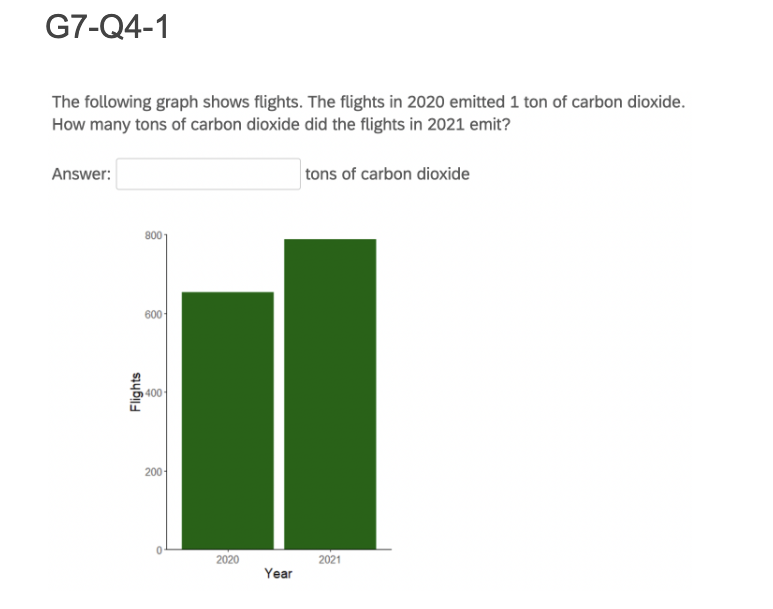} 
\end{figure}

\begin{figure}[htbp]
    \centering
    \includegraphics[width=0.48\textwidth]{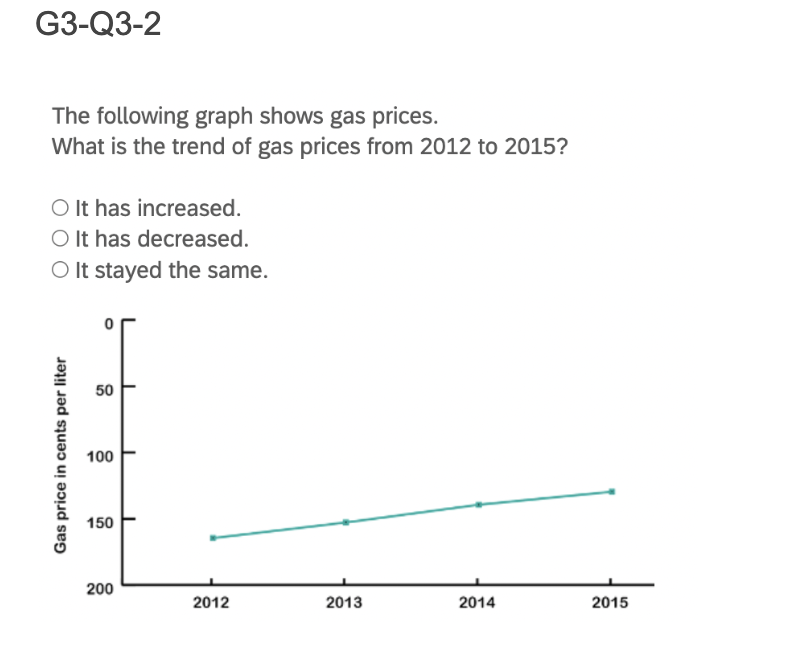} 
    \includegraphics[width=0.48\textwidth]{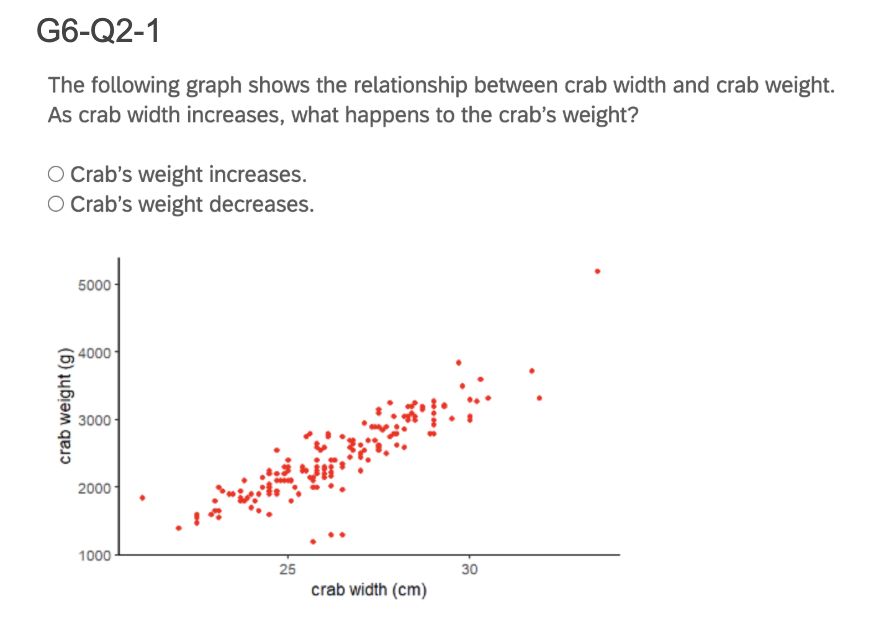} 
\end{figure}

Each participant received a \$20 Amazon gift card as compensation for their time and effort.

\newpage
\section*{Appendix B: Selected Charts in Table~\ref{tab:graph_types_scores}}

We present some selected chart and FACT question pairs from each manipulation group below.

\begin{figure}[htb]
\centering
\begin{tabular}{cc}

\includegraphics[width=0.48\textwidth]{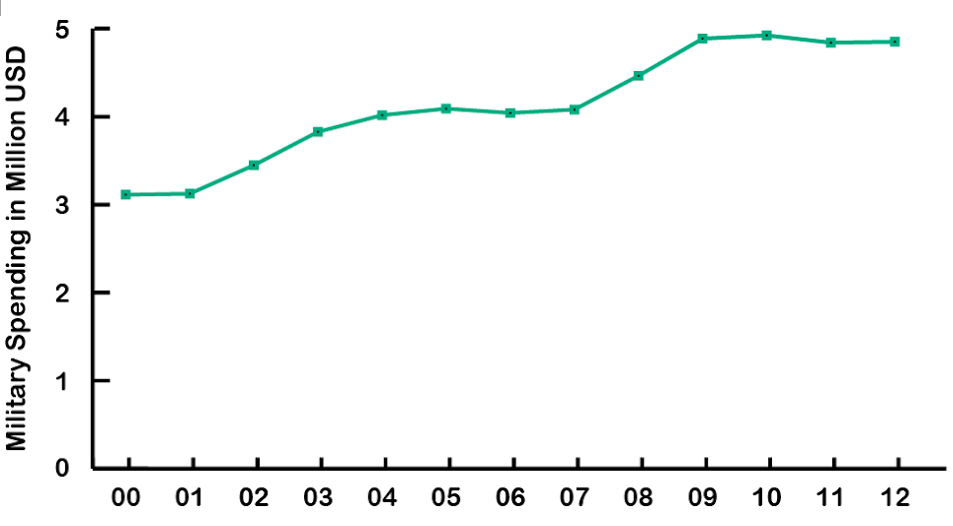} &
\includegraphics[width=0.48\textwidth]{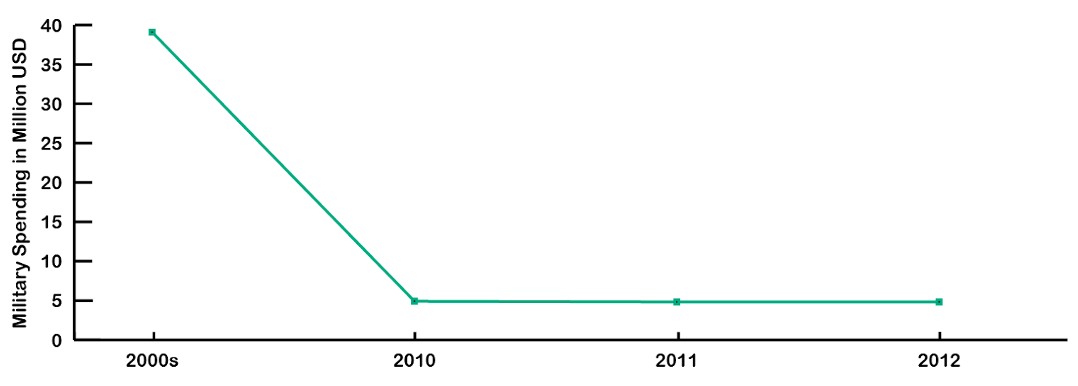} \\
G4\_Q1\_1 & G4\_Q1\_2 \\

\multicolumn{2}{l}{
\begin{minipage}{0.98\textwidth}
The following graph shows military spending. What is the trend of military spending from 2000 to 2012?
\\
1. It has increased.\quad 2. It has decreased.\quad 3. It stayed the same.
\end{minipage}}\\

\\
\hline 
\\

\includegraphics[width=0.48\textwidth]{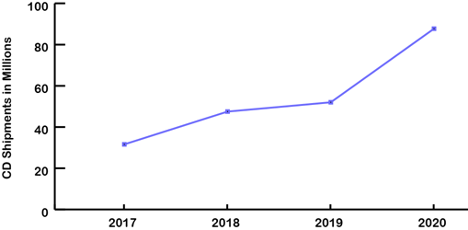} &
\includegraphics[width=0.48\textwidth]{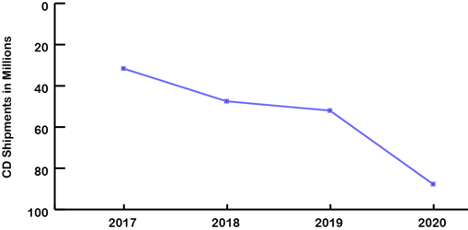} \\
G3\_Q1\_1 & G3\_Q1\_2 \\

\multicolumn{2}{l}{
\begin{minipage}{0.98\textwidth}
The following graph shows CD shipments. What is the trend of CD shipments from 2017 to 2020?\\
1. It has increased.\quad 2. It has decreased.\quad
3. It stayed the same.
\end{minipage}}\\

\\
\hline 
\\

\includegraphics[width=0.48\textwidth]{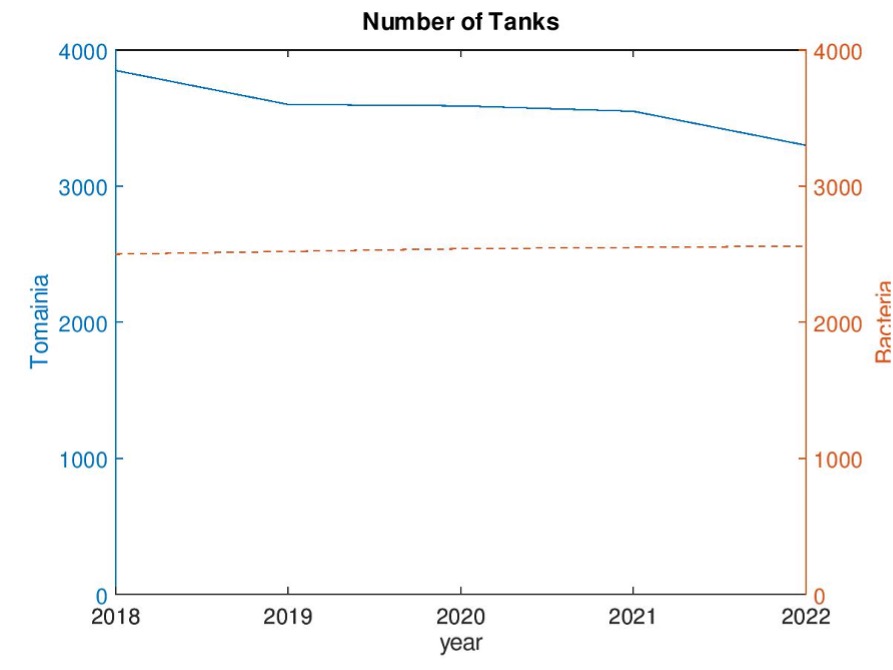} &
\includegraphics[width=0.48\textwidth]{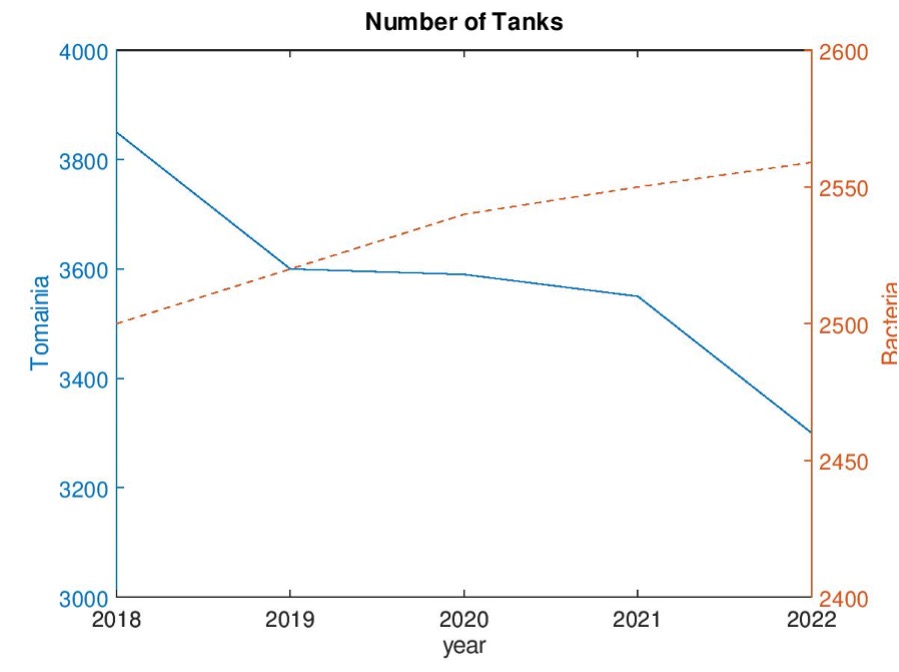} \\
G9\_Q1\_1 & G9\_Q1\_2 \\

\multicolumn{2}{l}{
\begin{minipage}{0.98\textwidth}
Two fictional countries Tomainia and Bacteria have been vying for military supremacy. The figure shows the number of tanks the countries had over the years. Which country had more tanks in 2021?
\\
1. Tomainia\quad 2. Bacteria
\end{minipage}}

\end{tabular}
\end{figure}

\begin{figure}[htb]
\centering
\begin{tabular}{cc}

\includegraphics[width=0.48\textwidth]{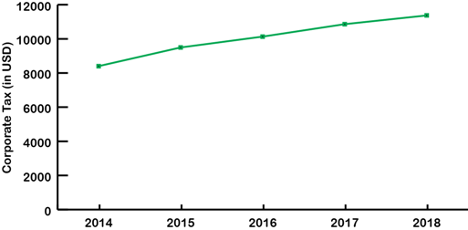} &
\includegraphics[width=0.48\textwidth]{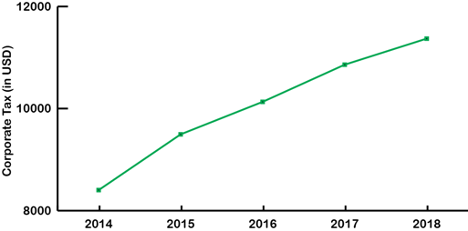} \\
G1\_Q1\_1 & G1\_Q1\_2 \\

\multicolumn{2}{l}{
\begin{minipage}{0.98\textwidth}
As a stack of \$1 bills, a company's corporate tax in 2014 was 1 yard high. How high is the company's corporate tax in 2018 as a stack of \$1 bills?
\end{minipage}}\\

\\
\hline 
\\

\includegraphics[width=0.48\textwidth]{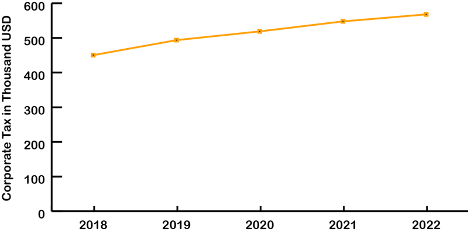} &
\includegraphics[width=0.48\textwidth]{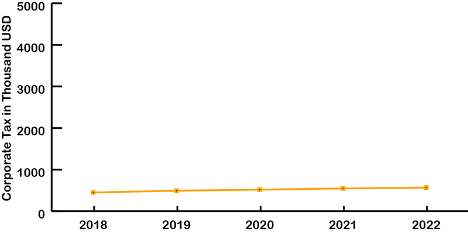} \\
G2\_Q1\_1 & G2\_Q1\_2 \\

\multicolumn{2}{l}{
\begin{minipage}{0.98\textwidth}
A company's corporate tax in 2018 weighed 1 ton in gold. How many tons in gold does the company's corporate tax weigh in 2022?
\end{minipage}}\\

\\
\hline 
\\

\includegraphics[width=0.48\textwidth]{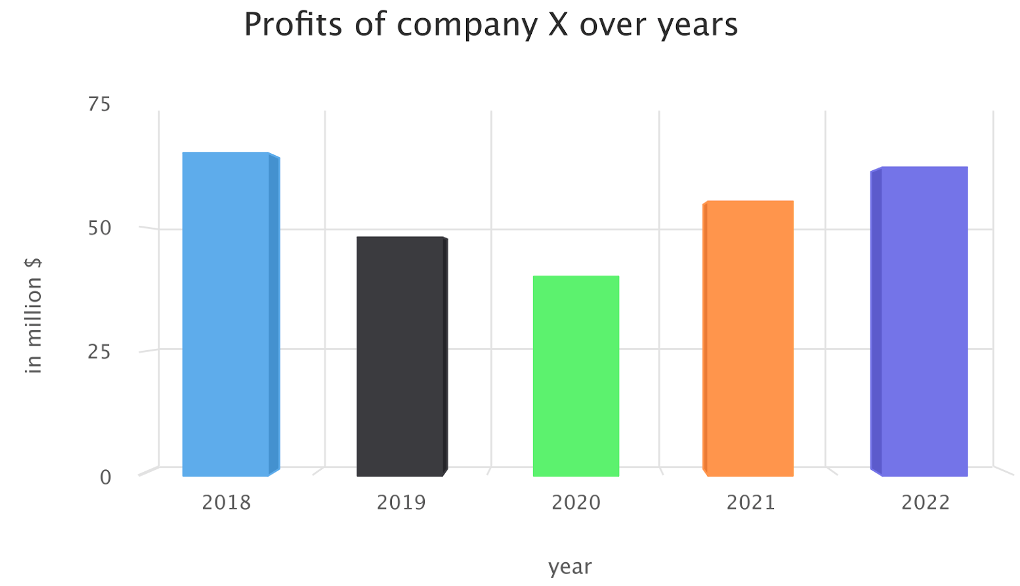} &
\includegraphics[width=0.48\textwidth]{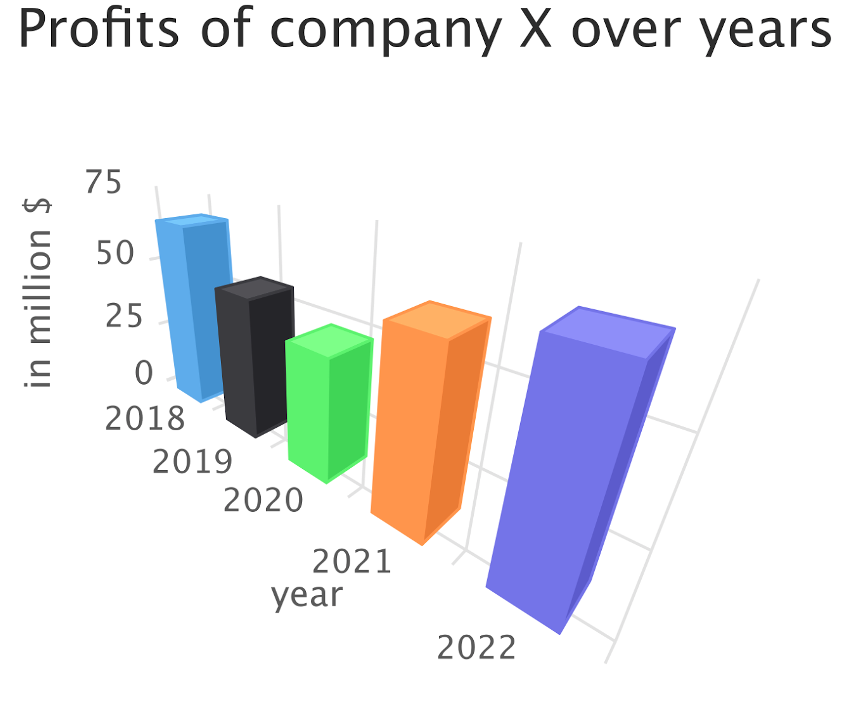} \\
G12\_Q1\_1 & G12\_Q1\_2 \\

\multicolumn{2}{l}{
\begin{minipage}{0.98\textwidth}
The following graph shows the profit of company X in millions. Which year was company X most profitable?
\\
1. 2018\quad 2. 2022\quad 3. 2018 and 2022 were the same.
\end{minipage}}\\

\end{tabular}
\end{figure}

\begin{figure}[htb]
\centering
\begin{tabular}{cc}

\includegraphics[width=0.35\textwidth]{graphics/G7_Q1_1.png} &
\includegraphics[width=0.35\textwidth]{graphics/G7_Q1_2.png} \\
G7\_Q1\_1 & G7\_Q1\_2 \\

\multicolumn{2}{l}{
\begin{minipage}{0.98\textwidth}
The following graph shows the number of dogs adopted. The dogs adopted in 2018 eat 1 million bags of dog food in their lifetimes. How much do the dogs adopted in 2019 eat in their lifetimes?
\end{minipage}}\\

\\
\hline 
\\

\includegraphics[width=0.35\textwidth]{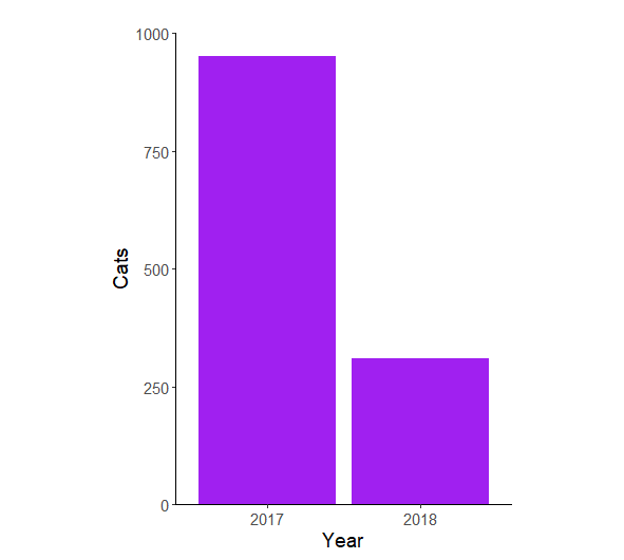} &
\includegraphics[width=0.35\textwidth]{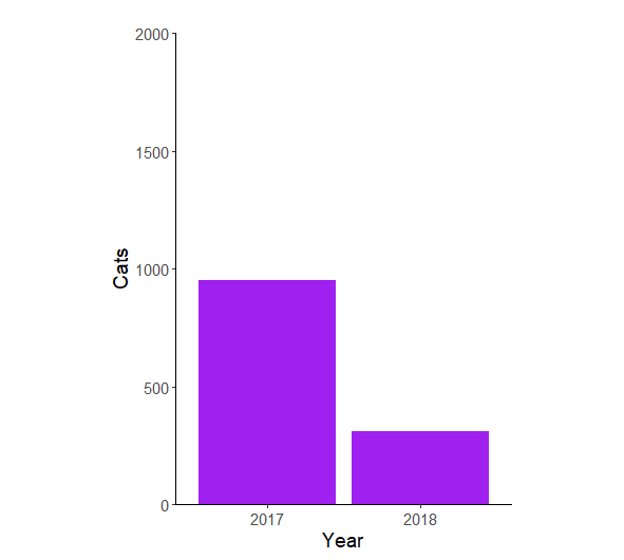} \\
G8\_Q1\_1 & G8\_Q1\_2 \\

\multicolumn{2}{l}{
\begin{minipage}{0.98\textwidth}
The following graph shows the number of cats adopted. The cats adopted in 2017 need 1 ton of kitty litter pellets. How many tons of kitty litter pellets do the cats adopted in 2018 need?
\end{minipage}}\\

\\
\hline 
\\

\includegraphics[width=0.4\textwidth]{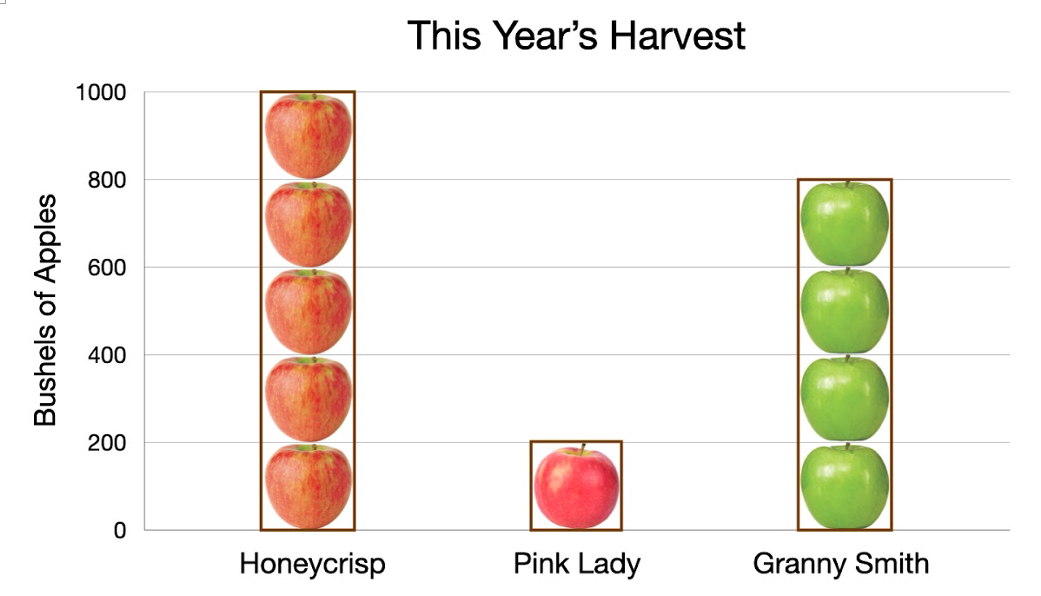} &
\includegraphics[width=0.4\textwidth]{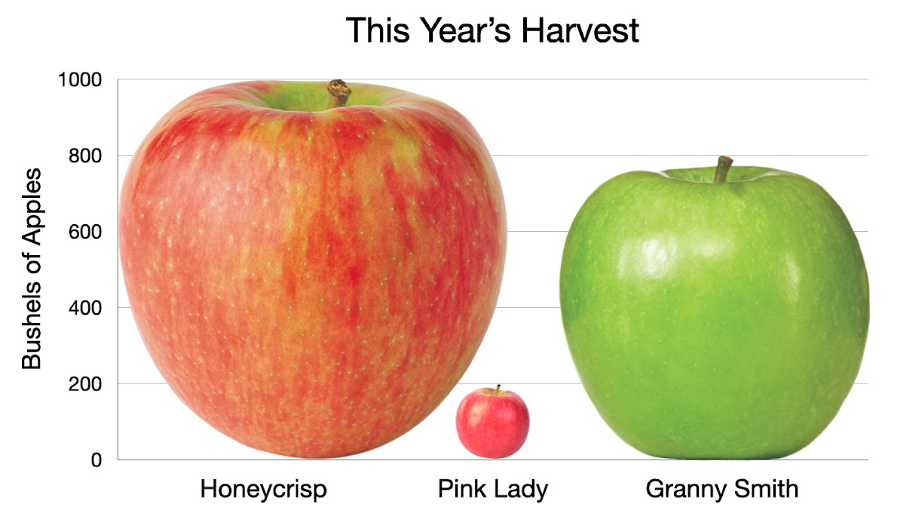} \\
G10\_Q1\_1 & G10\_Q1\_2 \\

\multicolumn{2}{l}{
\begin{minipage}{0.98\textwidth}
The graph shows how many bushels of apples were harvested this year for different kinds of apples. If 3 trucks are needed to transport all the Pink Lady apples, how many are needed to transport all the Honeycrisp apples?
\end{minipage}}\\

\end{tabular}
\end{figure}

\begin{figure}[htb]
\centering
\begin{tabular}{cc}

\includegraphics[width=0.4\textwidth]{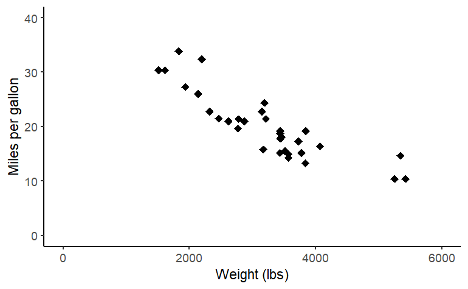} &
\includegraphics[width=0.4\textwidth]{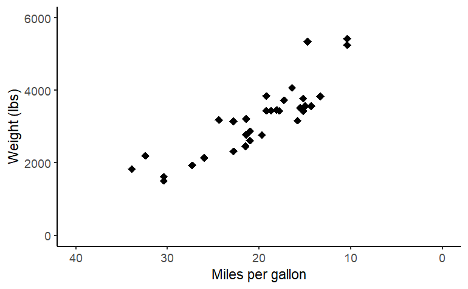} \\
G6\_Q1\_1 & G6\_Q1\_2 \\

\multicolumn{2}{l}{
\begin{minipage}{0.98\textwidth}
The following graph shows the relationship between car weight and miles per gallon. As car weight increases, what happens to the car's miles per gallon?
\\
1. Miles per gallon increases.\quad 
2. Miles per gallon decreases.
\end{minipage}}\\

\\
\hline 
\\

\includegraphics[width=0.4\textwidth]{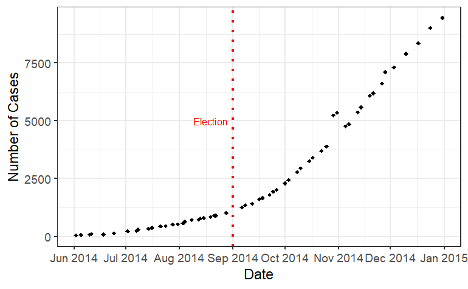} &
\includegraphics[width=0.4\textwidth]{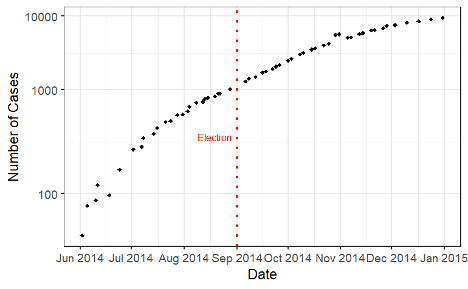} \\
G5\_Q1\_1 & G5\_Q1\_2 \\

\multicolumn{2}{l}{
\begin{minipage}{0.98\textwidth}
The following graph shows the number of ebola cases in Sierra Leone. An election took place in September 2014. When did ebola cases increase more?
\\
1. Before the election\quad 
2. After the election
\end{minipage}}\\

\\
\hline 
\\

\includegraphics[width=0.4\textwidth]{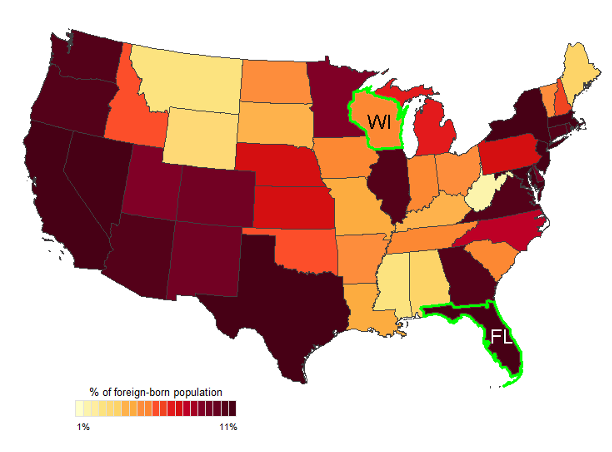} &
\includegraphics[width=0.4\textwidth]{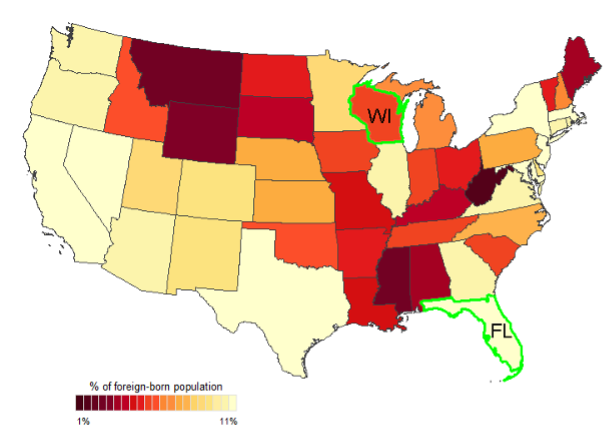} \\
G15\_Q1\_1 & G15\_Q1\_2 \\

\multicolumn{2}{l}{
\begin{minipage}{0.98\textwidth}
Which state has a higher foreign-born population, Wisconsin (WI) or Florida (FL)?
\\
1. Wisconsin (WI)\quad
2. Florida (FL)
\end{minipage}}\\

\end{tabular}
\end{figure}

\begin{figure}[htb]
\centering
\begin{tabular}{cc}

\includegraphics[width=0.48\textwidth]{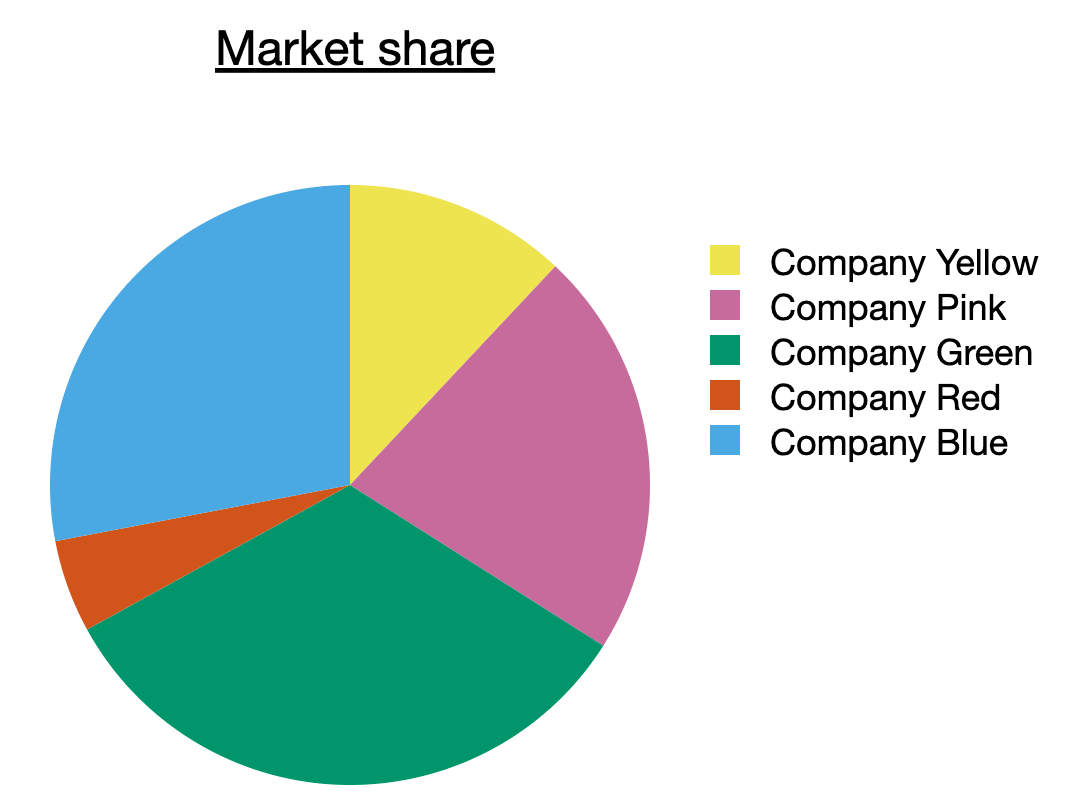} &
\includegraphics[width=0.48\textwidth]{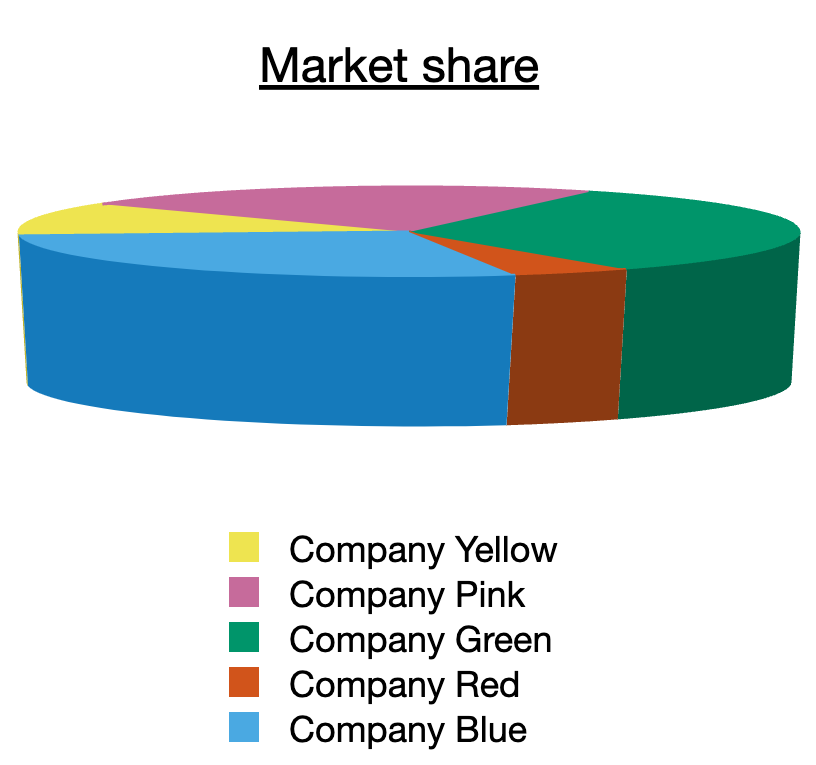} \\
G13\_Q1\_1 & G13\_Q1\_2 \\

\multicolumn{2}{l}{
\begin{minipage}{0.98\textwidth}
The figure below shows market share for different companies. Sort companies by market share from largest (1) to smallest (5).
The largest company (1) --- the smallest company (5)
\end{minipage}}\\

\\
\hline 
\\

\includegraphics[width=0.48\textwidth]{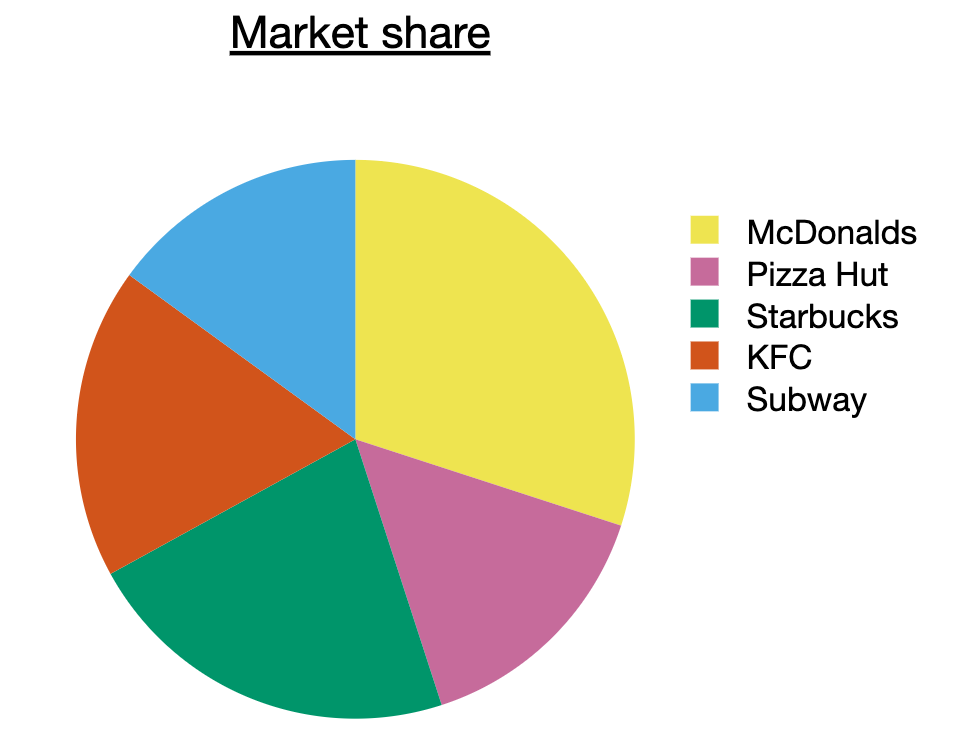} &
\includegraphics[width=0.48\textwidth]{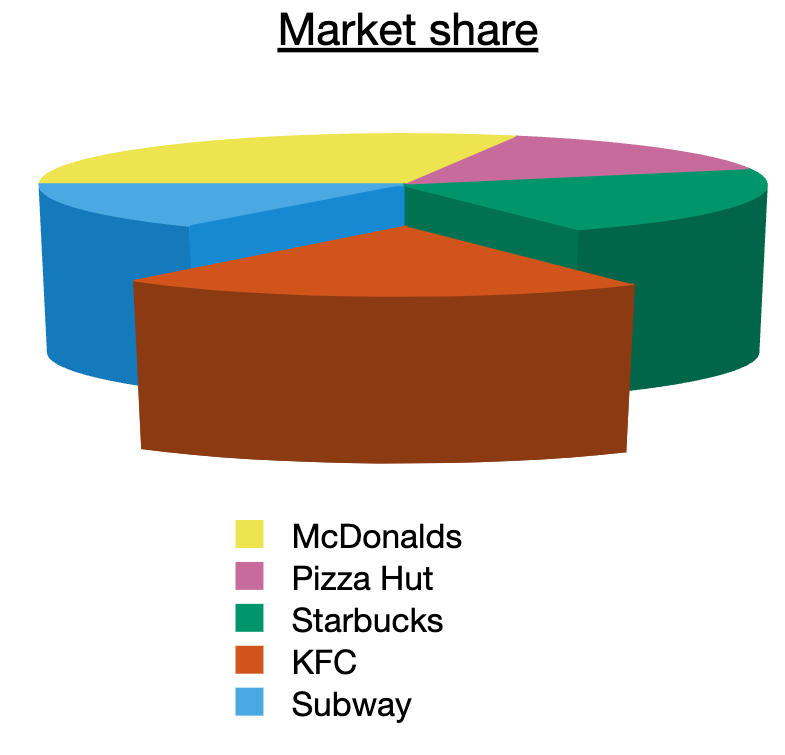} \\
G14\_Q1\_1 & G14\_Q1\_2 \\

\multicolumn{2}{l}{
\begin{minipage}{0.98\textwidth}
The figure below shows market share for different fast food restaurants. Based on this figure, sort restaurants by market share from largest (1) to smallest (5).
The largest company (1) --- the smallest company (5)
\end{minipage}}\\

\end{tabular}
\end{figure}